\documentclass[conference]{IEEEtran}
\IEEEoverridecommandlockouts
\usepackage{cite}
\usepackage{amsmath,amssymb,amsfonts}
\usepackage{algorithmic}
\usepackage{graphicx}
\usepackage{textcomp}
\usepackage{xcolor}
\usepackage{multirow}
\usepackage{bbm}
\usepackage{color,soul} 
\usepackage{colortbl}
\usepackage{enumitem}
\def\BibTeX{{\rm B\kern-.05em{\sc i\kern-.025em b}\kern-.08em
    T\kern-.1667em\lower.7ex\hbox{E}\kern-.125emX}}
\begin{document}
\newlength{\Oldarrayrulewidth}
\newcommand{\Cline}[2]{\noalign{\global\setlength{\Oldarrayrulewidth}{\arrayrulewidth}}\noalign{\global\setlength{\arrayrulewidth}{#1}}\cline{#2}\noalign{\global\setlength{\arrayrulewidth}{\Oldarrayrulewidth}}}

\title{SimCURL: Simple Contrastive User Representation Learning from Command Sequences}

\author{Hang Chu$^{1}$ \quad Amir Hosein Khasahmadi$^{1}$ \quad Karl D.D. Willis$^{1}$ \quad Fraser Anderson$^{2}$\\
Yaoli Mao$^{2}$ \quad Linh Tran$^{1}$ \quad Justin Matejka$^{2}$ \quad Jo Vermeulen$^{2}$\\
$^1$Autodesk AI Lab \quad $^2$Autodesk Research\\
{\tt\small \{hang.chu,amir.khasahmadi\}@autodesk.com}
}

\maketitle

\begin{abstract}
User modeling is crucial to understanding user behavior and essential for improving user experience and personalized recommendations.
When users interact with software, vast amounts of command sequences are generated through logging and analytics systems. 
These command sequences contain clues to the users' goals and intents.
However, these data modalities are highly unstructured and unlabeled, making it difficult for standard predictive systems to learn from.
We propose \textit{SimCURL}, a simple yet effective contrastive self-supervised deep learning framework that learns user representation from unlabeled command sequences.
Our method introduces a user-session network architecture, as well as session dropout as a novel way of data augmentation.
We train and evaluate our method on a real-world command sequence dataset of more than half a billion commands. 
Our method shows significant improvement over existing methods when the learned representation is transferred to downstream tasks such as experience and expertise classification.
\end{abstract}

\begin{IEEEkeywords}
user modeling, self-supervised learning
\end{IEEEkeywords}

\section{Introduction}
Accurate and robust user modeling has the potential to improve the user experience in a wide range of applications such as feature recommendation~\cite{matejka2009communitycommands,wang2018leveraging}, skill characterization~\cite{yang2018characterizing}, and software personalization~\cite{yang2017personalizing,tao2019log2intent}. 
Traditional predictive systems rely on annotated datasets. However, annotation is expensive and time-consuming.
In contrast, large amounts of raw command sequences can be passively stored without requiring human intervention.
We refer to a command sequence as a series of user-software interactions in the form of \textit{when} and \textit{what} commands are used.
Vast amounts of command sequences are generated via logging and analytics systems when users interact with software products and services, in both traditional desktop and modern cloud-based settings.
Using these command sequences offers new opportunities for user modeling by analyzing command sequences. 

Despite the availability of large amounts of data, its usability and application remain limited due to the lack of labels. Furthermore, acquiring labels is often difficult due to three main obstacles. First, annotating vast volumes of command data is costly and often impractical. Second, interpreting commands requires deep domain knowledge, which significantly increases labeling cost. Third, unlike natural language and images, the definition of common sense semantic knowledge is unclear for a command sequence. This causes each downstream user classification task to be separately labeled, with low label re-usability across different tasks. 

Self-supervised learning can be used to remove the dependency on massive amounts of labeled data in the conventional supervised learning paradigm. 
It is a great fit for command sequence representation learning, because it removes the bottleneck of labeled data. The backbone representation model can be directly trained on existing unlabeled command sequences.


\begin{figure}
\includegraphics[width=\columnwidth]{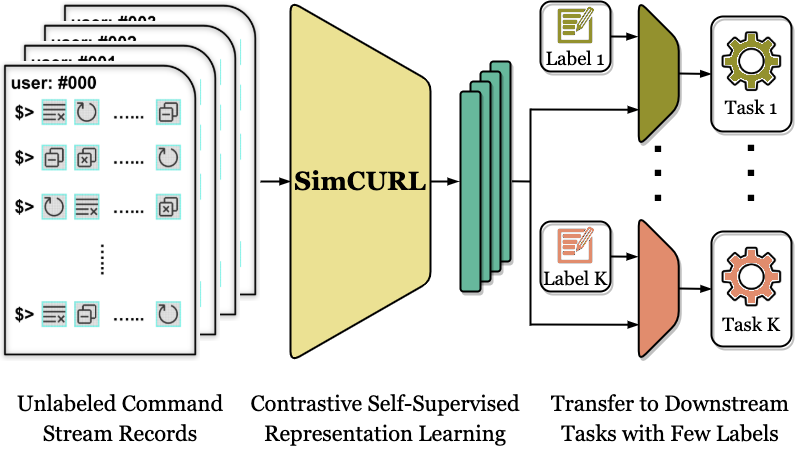}
\vspace{-8mm}
\caption{\textit{SimCURL} learns user representations from a large corpus of unlabeled command sequences. These learned representations are then transferred to multiple downstream tasks that have only limited labels available.}
\label{fig:teaser}
\end{figure}

We introduce \textit{SimCURL}, a Simple Contrastive User Representation Learning method that directly learns user embeddings from a large set of unlabeled command sequences (Figure~\ref{fig:teaser}).
\textit{SimCURL} takes a set of command sequences as input, where each command sequence is associated with a unique user identifier. Each element in the command sequence consists of a command name and a time tag. 
\textit{SimCURL} initially learns user representation vectors in a self-supervised manner. Then light-weight transfer learning models are trained with a small amount of labeled data for the downstream tasks.
To the best of our knowledge, \textit{SimCURL} is the first method to apply contrastive self-supervised representation learning to software command sequences.
Our contributions are as follows:

\begin{itemize}
\item We present a contrastive self-supervised learning method for learning user representations from unlabeled command sequences. 
\item We propose a new user-session neural network architecture, and a new session dropout data augmentation scheme to handle command sequences effectively.
\item We evaluate our method on a large-scale, real-world dataset with half billion commands and demonstrate significant improvement over existing methods.
\end{itemize}
\section{Related Work}

\begin{table}[t!]
\caption{Main statistics of the Fusion 360 command dataset.}
\vspace{-5mm}
\setlength{\tabcolsep}{14pt}
\begin{center}
\begin{tabular}{lcc}
\hline
~ & \textbf{Unlabeled Subset} & \textbf{Labeled Subset}\\ \hline
user & 199,996 & 12,612\\
session & 1,255,529 & 183,794\\
vocab & 3,164 & 3,164\\
command & 580,028,669 & 58,616,884\\ \hline
~ & \textbf{Task 1 (experience)} & \textbf{Task 2 (expertise)}\\ \hline
sub-task & 1 & 10\\
class & 8 & 3\\
user & 8,540 & 10,873\\ \hline
\end{tabular}
\end{center}
\label{tab:db_stat}
\vspace{-7mm}
\end{table}

\noindent \textbf{Self-Supervised Learning}: 
When conventional labels are not available, self-supervised learning can be used to leverage the data itself and its underlying structure as the source of the supervisory signal~\cite{lecun2021self}. 
It has a long history of successful deployments in the field of natural language processing such as the seminal work of word2vec~\cite{mikolov2013efficient} and GloVe~\cite{pennington2014glove}. 
Since the recent advances in training large models like Transformers~\cite{vaswani2017attention}, pre-trained language models such as BERT~\cite{devlin2018bert}, RoBERTa~\cite{liu2019roberta}, and T5~\cite{raffel2019exploring} have brought significant performance gains to a variety of natural language understanding tasks.
The GPT line of work~\cite{radford2018improving,radford2019language,brown2020language} utilizes the causal relationship of natural language tokens as the self-supervision objective. It is shown that the learned representation is multi-task transferable at both few-shot and zero-shot settings.

Self-supervised learning techniques have also been applied to audio~\cite{schneider2019wav2vec,niizumi2021byol}, image~\cite{he2020momentum,grill2020bootstrap,caron2021emerging}, and various other modalities~\cite{fang2021ssml,xue2021exploring}.
Due to the increasing difficulty of representing uncertainty in the prediction for images than it is for words~\cite{lecun2021self}, contrastive learning has emerged as an alternative self-supervised learning paradigm. The representation model is shown two distorted versions (or views) of the same data sample, and is trained to maximize their agreement.
The idea of learning representations by contrasting positive pairs and negative pairs dates back to Hadsell at el.~\cite{hadsell2006dimensionality}.
SimCLR~\cite{chen2020simple} simplifies the contrastive learning process by only using in-batch negative example sampling without resorting to a memory bank. 
SimCSE~\cite{gao2021simcse} applies the contrastive learning approach to natural language data and generates distorted views by simply dropping out words.
In this work, we take inspiration from SimCLR~\cite{chen2020simple} and SimCSE~\cite{gao2021simcse} due to their simplicity and effectiveness using in-batch negative examples. We perform data augmentation via the proposed session dropout technique in Section~\ref{sec:drop} .

\noindent \textbf{Sequence Modeling}: There is an extensive list of literature on sequence modeling techniques~\cite{viterbi1967error,rabiner1989tutorial,hochreiter1997long,chung2014empirical,vaswani2017attention}.
In this work, we treat command sessions of the same user as a temporal sequence and adopt the transformer architecture due to its ability to model long sequences and scale to large datasets.
A particularly relevant topic in sequence modeling is event modeling~\cite{mei2016neural}, where the sequence is not only temporally ordered, but also has associated time tags at each step.
Session-level modeling is an emerging technique in user behavior~\cite{wang2020beyond} and conversation~\cite{xiong2018session}, but has not yet been applied to command sequence data.
The main difference between event modeling and our work is granularity, i.e., event modeling focuses on modeling the sequence at event level, while our emphasis is learning user-level representations.

\noindent \textbf{Command Sequence Modeling}: User modeling and personalization are becoming the backbone of many software or social media businesses. 
Software platforms that record command sequences from consented users analyze these records to create user-specific recommendation and customization.
LogHub~\cite{he2020loghub} provides a collection of automatically generated logs from various systems and software.
However, these command stream collections are still rather small scale for training deep learning models.
CommunityCommands~\cite{matejka2009communitycommands} is amongst the early studies that tackle the task of recommending useful commands to the users. They analyzed the data of almost 40 million AutoCAD user command sequences for six months and proposed a measure coined CF-IUF, 
which is used for their collaborative filtering algorithm to provide personalized command recommendations.
In a more recent work~\cite{wang2018leveraging}, community-generated screen-recording videos are combined with command sequences to better classify and recommend software workflows. 
Gao et al.~\cite{gao2022command} addresses the problem of command prediction in CAD using transformers trained on augmented real-world data.

Util2vec~\cite{yang2017personalizing} presents a method that learns representations for both users and commands by predicting any command action given surrounding commands and the context vector of the user, which is used in user tagging and cold-start recommendation. 
Along the same line of research, log2int~\cite{tao2019log2intent} uses an auxiliary tutorial dataset~\cite{yang2019creative} for semantic encoding of the command sequences. Log2int introduces a deep sequence-to-sequence modeling approach to model the long-term temporal context in the command sequence.
Our method differs from util2vec and log2int in two ways: First, our representation learning objective is contrastive, instead of sequential modeling. Second, our method is better suited for learning user-level representations, instead of learning from local command sequence windows or sessions.
\section{Dataset}
\label{sec:data}

\begin{figure*}[t!]
\setlength{\tabcolsep}{5pt}
\renewcommand{\arraystretch}{2.5}
\begin{center}
\begin{tabular}{ccc}
\fbox{\includegraphics[width=0.3\textwidth]{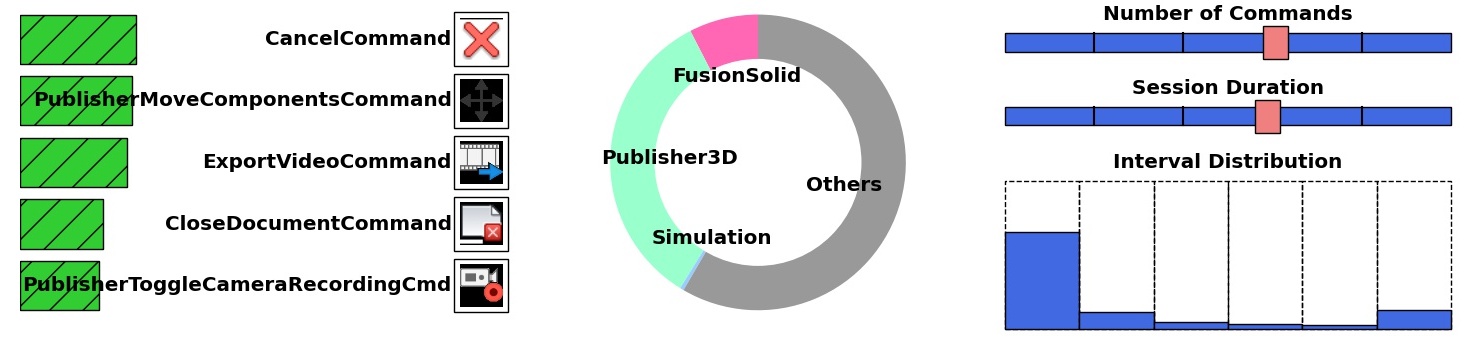}} & 
\fbox{\includegraphics[width=0.3\textwidth]{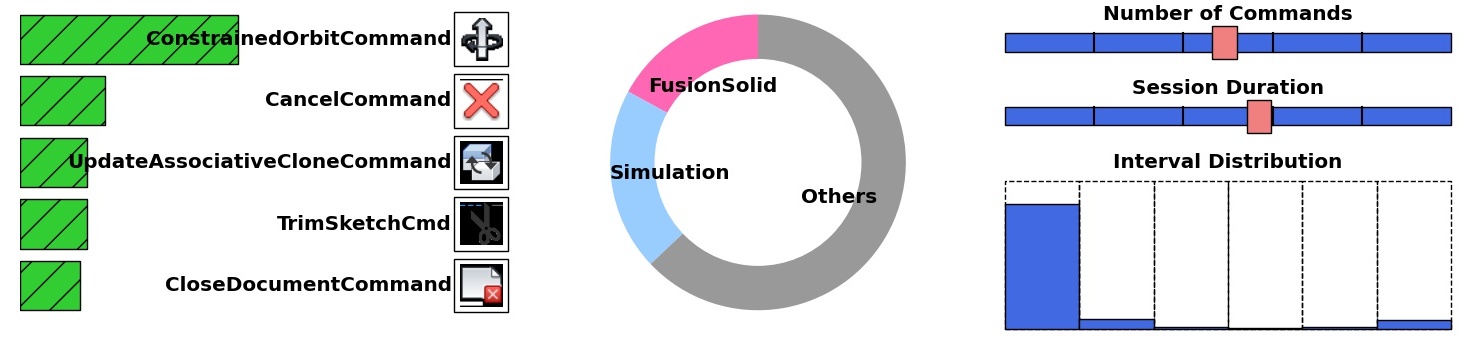}} & 
\fbox{\includegraphics[width=0.3\textwidth]{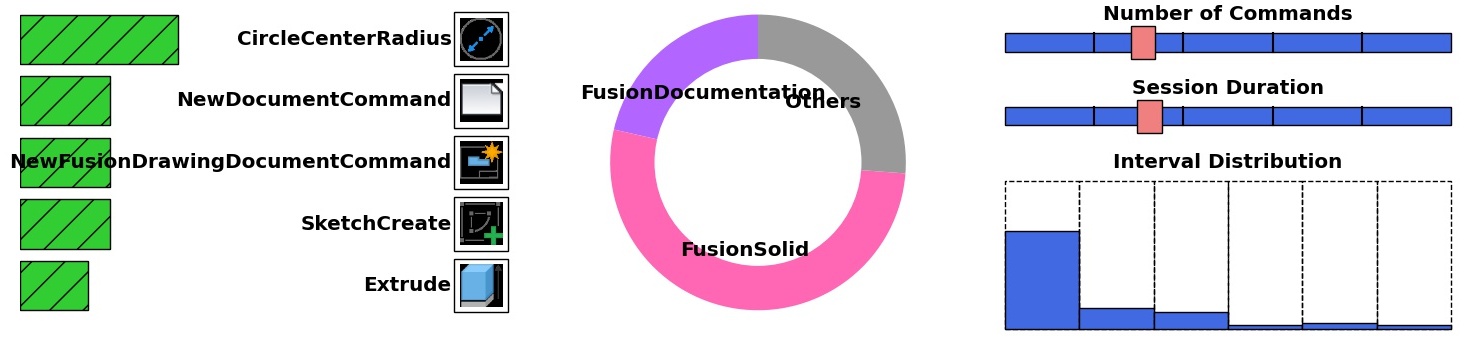}}\\
\fbox{\includegraphics[width=0.3\textwidth]{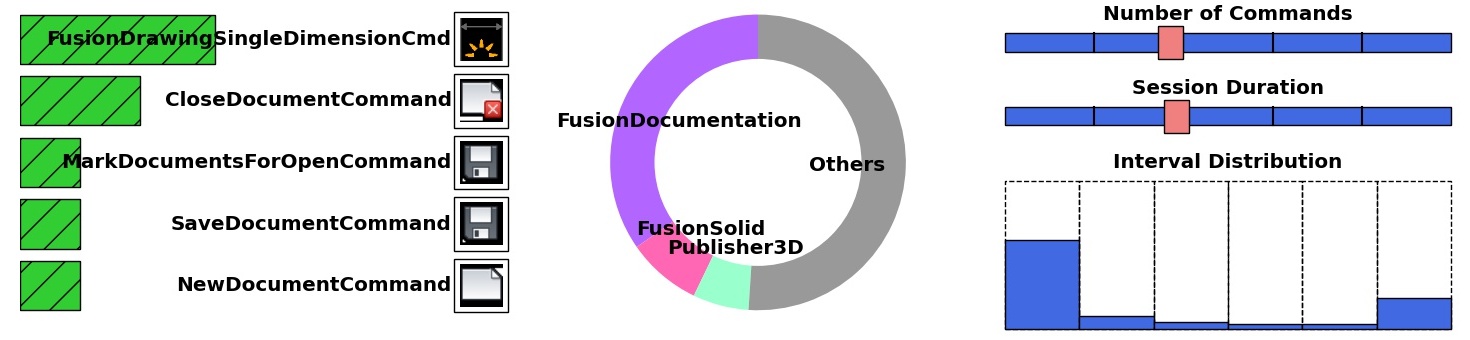}} & 
\fbox{\includegraphics[width=0.3\textwidth]{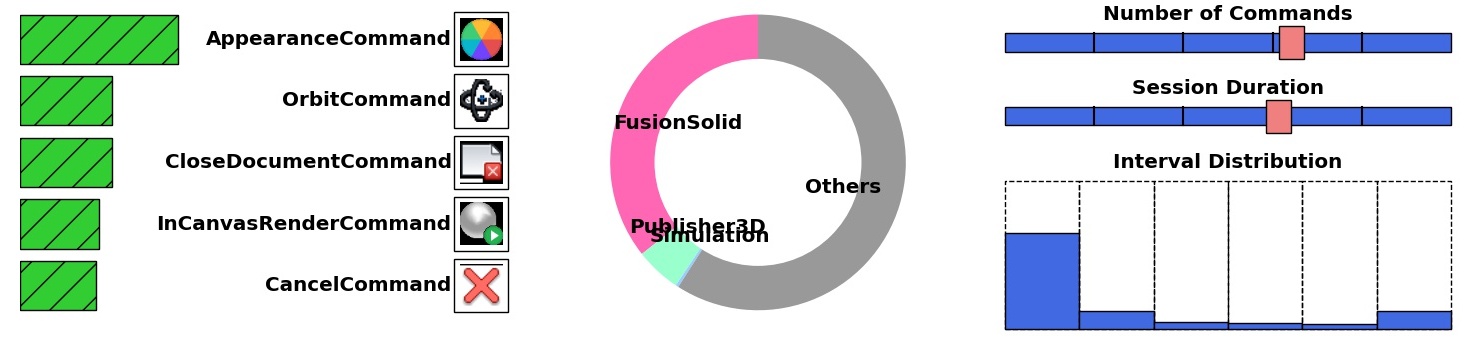}} & 
\fbox{\includegraphics[width=0.3\textwidth]{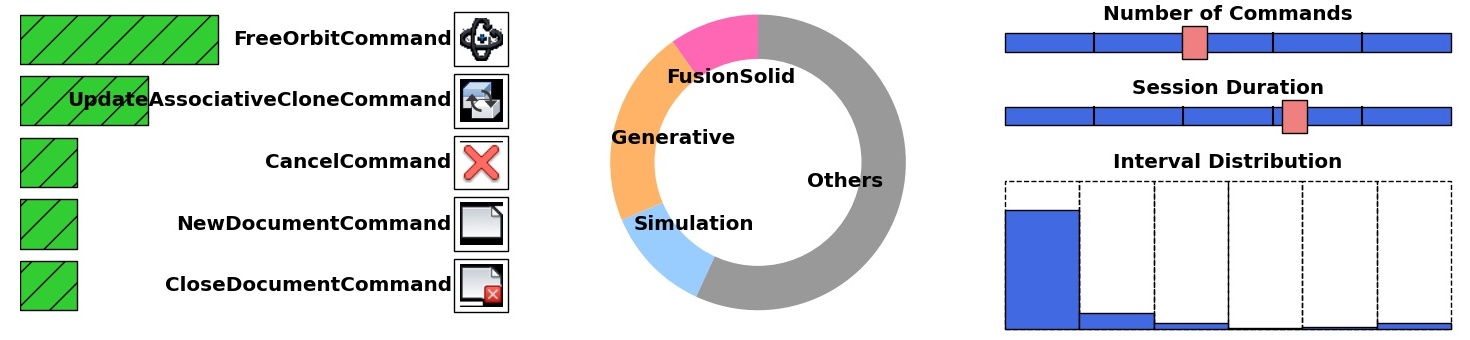}}\\
\end{tabular}
\end{center}
\vspace{-5mm}
\caption{Examples of Fusion 360 command sequence sessions in the dataset. For each example, the left column shows the top-5 most commonly used commands. The middle column shows the distribution of commands by their category. The right column shows the number of commands and session duration relatively to the overall dataset distribution, as well as the distribution of time interval between adjacent commands where the histogram bin width is 5 seconds.}
\label{fig:db_sample}
\vspace{-5mm}
\end{figure*}

Fusion 360\footnote{https://www.autodesk.ca/en/products/fusion-360/overview} is a commonly used cloud-connected desktop software from Autodesk that integrates Computer-Aided Design (CAD), Computer-Aided Manufacturing (CAM), Computer-Aided Engineering (CAE), and Printed Circuit Board (PCB) into one single software platform.
The multi-disciplinary nature and user diversity make Fusion 360 ideal for showcasing and benchmarking command sequence-based user representation learning.
To fully demonstrate the effectiveness of our self-supervised learning method, the dataset should satisfy two criteria: First, there should be a large-scale unlabeled set of raw command sequences. Second, there should be meaningful user classification downstream tasks with a small set of labels available, on which the learned representation can be evaluated. We now describe these two subsets of \textit{Unlabeled} and \textit{Labeled} data and provide their main statistics.

\subsection{Unlabeled Subset}

We collected command sequences of Fusion 360 users who have given us consent. We only recorded the name of the command, the time tag of each command, as well as an anonymous unique user identifier. For sophisticated software like Fusion 360, there are a great variety of commands (3164 unique commands). This means useful insight about the users' expertise and habits can be inferred based on what types of commands they use. At the same time, no information on the geometry or design parameters is collected to preserve the users' intellectual property.

We recorded command sequences within a 6 months period from October 2020 to April 2021.
We then randomly selected a subset of 199,996 users with at least 50 commands and at least two usage sessions within the 6 month period.
Table~\ref{tab:db_stat} shows the statistics of the unlabeled subset, which is large-scale, containing more than a million sessions and more than half a billion commands. 
Note that the categories are manually assigned by experts of the software and are only used for better illustration. 
Our method is general-purpose and does not depend on command category as its input.
Figure~\ref{fig:db_sample} shows common examples from the unlabeled subset. Note that the definition of sessions and how session boundaries are obtained will be described in the Section~\ref{sec:drop}. 
We observed that our dataset possesses a broad variety of different command sessions, e.g., a CAD solid modeling session that mainly contains solid modeling commands, a documentation session which involves documenting and viewing commands, or a mixed generative design and simulation session.

\subsection{Labeled Subset}
To evaluate our model and compare it with other baselines, we design several tasks that predict Fusion 360 user responses from a recent survey.
The Fusion 360 customer surveys are conducted annually to understand the product experience and inform future improvements. The 2020 survey contained 58 questions, covering customers' product usage, areas and levels of product expertise, and learning interests. 23,867 customers participated in the survey and opted in to the usage, processing, and storage of their data.

We use the data collected from the Fusion 360 customer survey to build a labeled subset. 
Specifically, we take two questions in the survey to make two user classification tasks. In the first task, we predict the user's years of experience in Fusion 360. For simplicity and keeping the task setting general purpose, we formulate this task as a 8-class classification, where each class represents a number of year ranging from 0 to 7. In the second task, we predict the areas of expertise of the user. This task contains 10 sub-tasks, each for one area of expertise, namely modeling, rendering, electronics, manufacturing CAM, extended manufacturing, animation, simulation, data collaboration, drawing, and generative technologies. Each sub-task is a 3-class classification task, indicating either the user has already used, or has not used, or plan to use the product features in the corresponding area of expertise. Similar to the unlabeled subset, we extract command sequences for users in the labeled subset. Table~\ref{tab:db_stat} lists the main statistics of the labeled subset. It can be seen that the labeled subset is approximately one order of magnitude smaller than the unlabeled subset. Note that because questions in the product survey are not always required, the number of users for the two tasks are different because not all survey participants have answered all questions.

Our selection of the two tasks in the labeled subset represents both challenging and relatively easy tasks. In Task 1, the model needs to possess a certain level of understanding about the user's proficiency, in order to better estimate years of experience based on only the most recent command stream records. Task 2 is less challenging than Task 1, because the types of commonly used commands correlate with a certain area of expertise. Despite this, it is still non-trivial to predict whether a user plans to use a new area of the product.
\section{Method}

\begin{figure*}[t!]
\begin{center}
\includegraphics[width=0.9\textwidth]{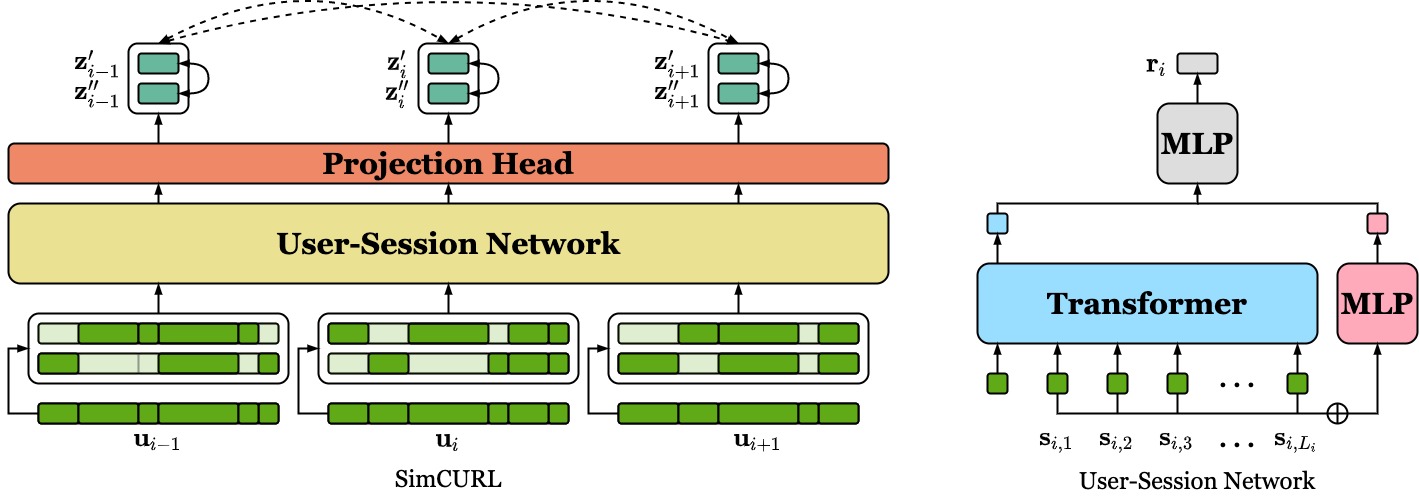}
\end{center}
\vspace{-7mm}
\caption{An overview of the \textit{SimCURL} method (left) and the user-session network architecture (right). The user's command sequence $\mathbf{u}_i$ is first divided into sessions $\{\mathbf{s}_{i,j}\}$, from which two augmented views are generated via session dropout. The views are passed through the main network to obtain the representation vectors $\mathbf{r}'_i$ and $\mathbf{r}''_i$, then the projection head to produce $\mathbf{z}'_{i}$ and $\mathbf{z}''_{i}$, on which the contrastive loss is applied. Solid and dashed lines denote positive and negative pairs, respectively.}
\label{fig:method}
\vspace{-5mm}
\end{figure*}

We define the command sequence dataset as a collection of user-specific command sequences $\mathcal{U}=\{\mathbf{u}_i\}$, where $1\leq i \leq N$ with $N$ being the number of users in the dataset.
For the $i$-th user, the corresponding $\mathbf{u}_i$ is a command sequence $\mathbf{u}_i=\{(c_{i,m},t_{i,m})\}$, where $1\leq m \leq M_i$ with $M_i$ denoting the total number of commands. At each command step $m$, the name of the command $c_{i,m}$ and the time tag of the command execution $t_{i,m}$ are recorded. 
Note that depending on the task and data availability, more command features can also be included besides the command name. In this work, we only use the categorical $c_{i,m}$ for simplicity without loss of generality.

In the conventional supervised learning setting, each user is labeled as $y_i$ in a given downstream task, and the training objective function for fully supervised learning can be written as:
\begin{align}
\mathbf{\theta}_f = \underset{\mathbf{\theta}_f}{\mathrm{argmin}}-\sum^{N}_{i=1} \mathrm{log}\big(f\left(y_i | \mathbf{u}_i\right)\big)
\end{align}
where $f$ is the model that predicts class probabilities given the user's command sequence, and $\mathbf{\theta}_f$ denotes its trainable parameters. In the self-supervised representation learning setting, we denote $\mathcal{U}^{\alpha}=\{\mathbf{u}^{\alpha}_i\}$ as the unlabeled self-supervised learning subset, and $\mathcal{U}^{\beta}=\{\mathbf{u}^{\beta}_i\}$ as the labeled supervised learning subset labeled as $y^{\beta}_{i}$, with sizes of $N^{\alpha}$ and $N^{\beta}$, respectively. The self-supervised learning training objective becomes:
\begin{align}
\mathbf{\theta}_g = \underset{\mathbf{\theta}_g}{\mathrm{argmin}}\sum^{N^{\alpha}}_{i=1}\mathcal{L}\big(g\left(\mathbf{u}_{i}^{\alpha}\right)\big)
\end{align}
where $g$ is the main model that produces the representation vectors. $\mathbf{\theta}_g$ denotes its trainable parameters and $\mathcal{L}$ is the self-supervised learning contrastive loss function. After the first stage which trains the main model, the learned representation is transferred to a downstream task, whose training objective is:
\begin{align}
\mathbf{\theta}_f = \underset{\mathbf{\theta}_f}{\mathrm{argmin}}-\sum^{N^{\beta}}_{i=1} \mathrm{log}\Big(f\big(y_{i}^{\beta} | g(\mathbf{u}_{i}^{\beta})\big)\Big)
\end{align}
which follows the same formulation as the supervised learning setting, except that users are first encoded via $g$.
It should be noted that there can be multiple downstream tasks. In that case, a different classification model $f_k$ is trained for each task, but the main model $g$ stays the same. We omit the subscript $k$ to keep the notation simple.

Our proposed method contains three main steps: session segmentation and data augmentation via session dropout, the user-session neural network architecture, and the final contrastive self-supervised representation learning objective. Next, we describe these main components in detail respectively.

\subsection{Session Dropout}
\label{sec:drop}
Different from the conventional definition where a session can be defined as a segment of commands between a log-in and a log-out action, we define a command session as a contiguous temporal segment of commands. This is out of two main reasons: First, there may not be clearly defined session boundary commands in some software. Second, a user could keep the software running without closing the window for a long time due to frequent usage.
We compute the grouping of commands into sessions by first finding relative peaks of command density over time, then clustering commands into the nearest peak to form the sessions.

Concretely, we first discretize the time axis into $Q$ equidistant bins, with the time value at bin centers as $\{t_q\}$ where $1\leq q \leq Q$. Then for each bin of the $i$-th user, we compute its command execution density $d_{i,q}$ as:
\begin{align}
d_{i,q} = \sum_{m=1}^{M}\mathrm{exp}\bigg(-\Big(\frac{t_q-t_{i,m}}{\sigma}\Big)\bigg)
\end{align}
where $\sigma$ is a constant number that controls the amount of temporal smoothing. We then find the set of local maximum peaks $\mathbf{p}_i=\{p_{i,j}\}$ amongst $\{d_{i,q}\}$ as:
\begin{align}
\mathbf{p}_i = \big\{t_q\;\big|\;d_{i,q}\geq d_{i,q+\Delta},\forall \Delta \in [-w,w]\big\}
\end{align}
where $w$ denotes a constant window size. Finally, the commands are grouped into sessions $\{\mathbf{s}_{i,j}\}$ based on their nearest peaks as:
\begin{align}
\mathbf{s}_{i,j}=\big\{(c_{i,m},t_{i,m})\;\big|\;|t_{i,m}-p_{i,j}|\leq|t_{i,m}-p_{i,x}|,\forall p_{i,x} \in \mathbf{p}_i\big\}
\end{align}
where the number of sessions is equal to the number of peaks, which we denote as $L_i$. Thus, the original single command stream is partitioned into sessions of 
$\mathbf{u}_i=\{\mathbf{s}_{i,j}\}$, where $1\leq j \leq L_i$.

The effectiveness of contrastive representation learning relies on data augmentation techniques that apply random distortion, but still preserve unique characteristics of the data point.
To achieve this, we simply apply dropout at the session level. By using session dropout, we can not only produce large variety of views for a single data point, but also maintain the completeness of the user's workflow within sessions.

\subsection{User-Session Network}
We propose a new user-session network architecture to encode command stream data effectively. There are three main factors that motivate our network design:
First, despite recent advancements in sequence modeling techniques, it is impractical to directly feed the command sequence into the model because the raw sequence length $M$ is typically long. This is evident in Table~\ref{tab:db_stat}, where the average length is 2900 for the time span over six months. Also to achieve the goal of user modeling, it is less important to reason at the command-level granularity. 
Second, it is useful to take session-level information into account. As can be seen in Figure~\ref{fig:db_sample} there is a wide diversity of sessions. Such diversity is also exhibited within the same user, where different sessions tackle different projects or different stages of the same project. Therefore, we set session as the time step unit for our model.
Third, we found it helpful to directly encode user-level information with a dedicated branch, which aligns with the objective of learning user-level representations.

The right-hand side of Figure~\ref{fig:method} shows the architecture of our user-session network.
For each session $\mathbf{s}_{i,j}$, we simply use a normalized vector of command usage frequency as its input feature. Similarly, at the user-level we use the overall command usage frequency vector as the input feature for the user branch.
We use a Transformer network~\cite{vaswani2017attention} to encode the sequence of sessions. At each time step, the input feature vector is first passed through a linear layer. Then, it is added to the position encoding vector that encodes the time step via a series of sinusoidal functions, and passed through a stack of multi-head self-attention layers.
We also reserve the first time step for a <rep> token that does not involve any input session. Since in the Transformer, the query-key-value computation is conducted on every pairs of time steps, it is sufficient to directly use the Transformer output at the first time step as the session-level encoding.
Finally, we use a simple Multi-Layer Perceptron (MLP) network to encode the user-level information, as well as another MLP that combines the user-level and session-level encodings to produce the user representation vector $\mathbf{r}_i$.

\subsection{Contrastive Loss}
\textit{SimCURL} learns self-supervised user representations by applying a contrastive loss term on in-batch positive and negative examples.
To achieve this, each user is transformed into two differently augmented views using session dropout described above, then fed into the user-session network $g$. Following the common practice in contrastive self-supervised learning, we also use another multi-layer projection head denoted as $h$ consisting of linear layers. We obtain the final vectors as:
\begin{align}
\mathbf{z}'_i = h\Big(g\big(sd'(\mathbf{u}_i)\big)\Big),\;
\mathbf{z}''_i = h\Big(g\big(sd''(\mathbf{u}_i)\big)\Big)
\end{align}
where $sd'$ and $sd''$ denotes the session dropout function with two different random seeds. As depicted in the left-hand side of Figure~\ref{fig:method}, the two augmented views of each command sequence form the positive examples, while all other pairs form the negative examples. The final contrastive loss can be written as:
\begin{gather}
\mathcal{L} = \sum_{i}^{B}\big(\mathcal{L}'_{i}+\mathcal{L}''_{i}\big)\\
\mathcal{L}'_{i} = -\mathrm{log}\frac{\mathrm{exp}\Big(\mathrm{sim}\big(\mathbf{z}'_i,\mathbf{z}''_i\big)\Big)}{\sum_{i'\neq i}^{B}\mathrm{exp}\Big(\mathrm{sim}\big(\mathbf{z}'_{i'},\mathbf{z}'_{i}\big)\Big) + \sum_{i''\neq i}^{B}\mathrm{exp}\Big(\mathrm{sim}\big(\mathbf{z}''_{i''},\mathbf{z}'_{i}\big)\Big)}\\
\mathrm{sim}\big(\mathbf{z}'_i,\mathbf{z}''_i\big) = \frac{\mathbf{z}'_i\cdot \mathbf{z}''_i}{\tau\: ||\mathbf{z}'_i||\: ||\mathbf{z}''_i||}
\end{gather}
where $B$ denotes the mini-batch size and indicates the loss is computed over samples within the same training mini-batches, $\mathrm{sim(\cdot)}$ denotes the cosine similarity function with an extra constant temperature parameter denoted as $\tau$. $\mathcal{L}''$ is similarly defined as $\mathcal{L}'$ for the second view instead of the first view.  
Intuitively, the loss function pulls the positive examples closer while pushing the negative examples further in the latent space. During training, we randomly sample mini-batches of users, minimize the loss for the sampled mini-batch, and repeat this process until convergence is reached. We optimize the parameters $\mathbf{\theta}_{g}$ and $\mathbf{\theta}_{h}$ for both the main user-session network $g$ and the projection head $h$ during training, and throw away $h$ after training is completed.

\subsection{Implementation Details}
For session division, we empirically set the temporal quantization level $Q$ as $2^{15}$, and the temporal smoothing constant $\sigma$ as $2^{10}$. We set the window size for command density peak detection $w$ as $1$. In the contrastive loss, we simply set the temperature constant $\tau$ as $1$. We use the ReLU activation function in the MLP modules and the projection head. $4$ attention heads are used in the session Transformer module. Hidden vector sizes are consistently set as $256$ for both $\mathbf{r}_i$ and $\mathbf{z}_i$. The model is trained with the Adam~\cite{kingma2014adam} optimizer on a Nvidia RTX6000 GPU with a 24G memory for 40 epochs.
After the contrastive self-supervised learning is completed on the unlabeled subset $\mathcal{U}^{\alpha}$, we train a linear classifier as the transfer learning model $f$ on top of the user-session network $g$ whose weights are set frozen. We also apply session dropout on the users $\mathbf{u}_i^{\beta}$ during the downstream task training stage. 

\section{Experiments}

\begin{table*}[t!]
\caption{Main results using full labeled subsets. {\color{green}Best} and {\color{blue}second best} results are highlighted in {\color{green} green} and {\color{blue} blue}.}
\vspace{-3mm}
\setlength{\tabcolsep}{6pt}
\begin{tabular}{l|cc|cc|ccc}
\hline
\multirow{2}{*}{~} & \multicolumn{2}{c|}{Task 1 (experience)} & \multicolumn{2}{c|}{Task 2 (expertise)} & \multicolumn{3}{c}{Overall}\\ \cline{2-8}
& Acc./\% & F1/\% & Acc./\% & F1/\% & Acc./\% & F1/\% & ROC-AUC\\ \hline
random & $13.29 \pm 0.92$ & $14.50 \pm 0.49$ & $33.17 \pm 0.16$ & $38.32 \pm 0.04$ & $23.23 \pm 0.54$ & $26.41 \pm 0.26$ & \multirow{8}{*}{\includegraphics[width=0.29\columnwidth]{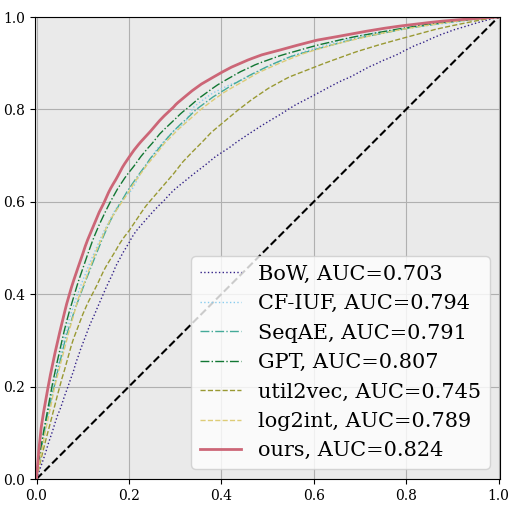}}\\ \cline{1-7}
BoW & $28.14 \pm 0.71$ & $27.40 \pm 0.74$ & $57.62 \pm 0.15$ & $56.98 \pm 0.51$  & $42.88 \pm 0.32$ & $42.19 \pm 0.41$\\ \cline{1-7}
CF-IUF~\cite{matejka2009communitycommands} & $31.50 \pm 0.11$ & $19.17 \pm 0.30$ & $67.67 \pm 0.01$ & $55.98 \pm 0.02$  & $49.58 \pm 0.05$ & $37.58 \pm 0.15$\\ \cline{1-7}
SeqAE~\cite{dai2015semi} & $31.90 \pm 0.13$ & $29.05 \pm 1.26$ & $65.82 \pm 0.65$ & \cellcolor{blue!25}$58.71 \pm 0.11$  & $48.86 \pm 0.30$ & $43.88 \pm 0.68$\\ \cline{1-7}
GPT~\cite{radford2018improving} & \cellcolor{blue!25}$33.44 \pm 1.20$ & \cellcolor{blue!25}$30.81 \pm 0.86$ & \cellcolor{blue!25}$67.81 \pm 0.20$ & $58.70 \pm 0.70$  & \cellcolor{blue!25}$50.62 \pm 0.52$ & \cellcolor{blue!25}$44.76 \pm 0.55$\\ \cline{1-7}
util2vec~\cite{yang2017personalizing} & $25.68 \pm 0.53$ & $24.55 \pm 0.53$ & $64.52 \pm 0.26$ & $57.45 \pm 0.90$  & $45.10 \pm 0.36$ & $41.00 \pm 0.25$\\ \cline{1-7}
log2int~\cite{tao2019log2intent} & $31.78 \pm 0.63$ & $29.54 \pm 0.17$ & $64.82 \pm 0.20$ & $58.64 \pm 0.31$  & $48.30 \pm 0.22$ & $44.09 \pm 0.16$\\ \cline{1-7} \cline{1-7}
ours & \cellcolor{green!25}$\mathbf{37.39 \pm 0.41}$ & \cellcolor{green!25}$\mathbf{35.24 \pm 0.68}$ & \cellcolor{green!25}$\mathbf{68.67 \pm 0.08}$ & \cellcolor{green!25}$\mathbf{60.26 \pm 0.07}$ & \cellcolor{green!25}$\mathbf{53.03 \pm 0.22}$ & \cellcolor{green!25}$\mathbf{47.75 \pm 0.37}$\\ \hline
\end{tabular}
\label{tab:main}
\vspace{-10pt}
\end{table*}

\begin{table*}[t!]
\caption{Few-shot learning that trains with only 6.25\% labels. {\color{green}Best} and {\color{blue}second best} results are highlighted in {\color{green} green} and {\color{blue} blue}.}
\vspace{-3mm}
\setlength{\tabcolsep}{6pt}
\begin{tabular}{l|cc|cc|ccc}
\hline
\multirow{2}{*}{~} & \multicolumn{2}{c|}{Task 1 (experience)} & \multicolumn{2}{c|}{Task 2 (expertise)} & \multicolumn{3}{c}{Overall}\\ \cline{2-8}
& Acc./\% & F1/\% & Acc./\% & F1/\% & Acc./\% & F1/\% & ROC-AUC\\ \hline
random & $12.16 \pm 0.09$ & $14.67 \pm 0.66$ & $33.50 \pm 0.30$ & $38.40 \pm 0.37$ & $22.83 \pm 0.18$ & $26.53 \pm 0.49$ & \multirow{8}{*}{\includegraphics[width=0.29\columnwidth]{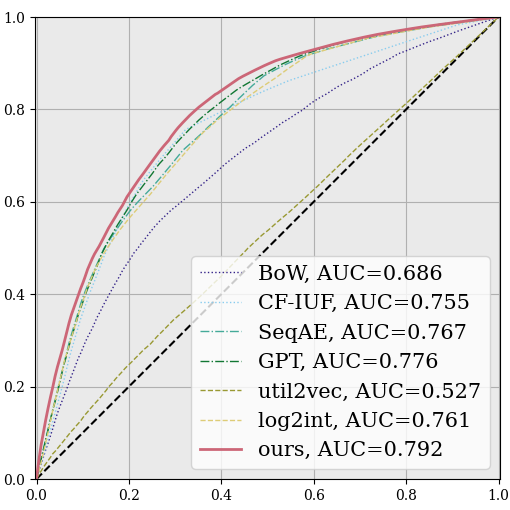}}\\ \cline{1-7}
BoW & $25.12 \pm 0.67$ & \cellcolor{blue!25}$24.74 \pm 0.62$ & $55.23 \pm 1.13$ & $55.41 \pm 0.75$  & $40.18 \pm 0.42$ & $40.07 \pm 0.08$\\ \cline{1-7}
CF-IUF~\cite{matejka2009communitycommands} & \cellcolor{blue!25}$29.09 \pm 1.77$ & $19.79 \pm 2.77$ & \cellcolor{green!25}$\mathbf{67.27 \pm 0.04}$ & $54.86 \pm 0.13$  & \cellcolor{blue!25}$48.18 \pm 0.88$ & $37.33 \pm 1.33$\\ \cline{1-7}
SeqAE~\cite{dai2015semi} & $28.11 \pm 1.48$ & $17.45 \pm 0.98$ & $65.06 \pm 1.87$ & $55.42 \pm 0.43$  & $46.59 \pm 1.53$ & $36.43 \pm 0.39$\\ \cline{1-7}
GPT~\cite{radford2018improving} & $28.81 \pm 1.09$ & $23.89 \pm 1.64$ & $66.41 \pm 0.19$ & \cellcolor{blue!25}$57.75 \pm 0.27$  & $47.61 \pm 0.55$ & \cellcolor{blue!25}$40.82 \pm 0.95$\\ \cline{1-7}
util2vec~\cite{yang2017personalizing} & $14.54 \pm 0.78$ & $16.74 \pm 1.01$ & $37.00 \pm 0.32$ & $41.90 \pm 0.31$  & $25.77 \pm 0.50$ & $29.32 \pm 0.60$\\ \cline{1-7}
log2int~\cite{tao2019log2intent} & $25.21 \pm 4.42$ & $14.15 \pm 5.06$ & $64.73 \pm 1.57$ & $54.92 \pm 1.25$  & $44.97 \pm 2.99$ & $34.53 \pm 3.11$\\ \cline{1-7} \cline{1-7}
ours & \cellcolor{green!25}$\mathbf{33.32 \pm 0.67}$ & \cellcolor{green!25}$\mathbf{30.49 \pm 1.01}$ & \cellcolor{blue!25}$67.20 \pm 0.12$ & \cellcolor{green!25}$\mathbf{59.60 \pm 0.27}$ & \cellcolor{green!25}$\mathbf{50.26 \pm 0.35}$ & \cellcolor{green!25}$\mathbf{45.04 \pm 0.39}$\\ \hline
\end{tabular}
\label{tab:fewshot}
\vspace{-5mm}
\end{table*}

We show experimental results of the proposed \textit{SimCURL} method in this section, and compare them with baselines as well as methods adapted from previous literature. We first describe our experimental settings, followed by detailed discussions on the performance comparisons.
We split the unlabeled subset by the number of users at a 8-1-1 ratio into the training, validation, and testing sets. For the downstream tasks, we split the data evenly by the number of users at a 1-1-1 ratio. To evaluate the performance of our proposed method, we compare against a random baseline as well as six more sophisticated methods:
\begin{itemize}[leftmargin=*]
\item \textbf{BoW}: A Bag-of-Words baseline that directly represents the user as the frequency of commands that has appeared in the user's command sequence.
\item \textbf{CF-IUF~\cite{matejka2009communitycommands}}: Another baseline representation that further enhances BoW by representing the user using a vector by taking the element-wise product between the command frequency vector and a logarithmic scale inverse user frequency vector. The user frequency vector further provides information about the commands' popularity amongst all users, indicating the total percentage of users that have used each command in the vocabulary.
\item \textbf{SeqAE~\cite{dai2015semi}}: A baseline that uses a sequential auto-encoder inspired by Dai et al.~\cite{dai2015semi}. We use two recurrent neural networks with GRU layers. The first network encodes the command sequence of a session, and the second decoder network takes the encoding vector and tries to reconstruct the session. We only keep the encoder network that produces the user representation, and use it to train linear classifiers in the downstream tasks.
\item \textbf{GPT~\cite{radford2018improving}}: A baseline that uses an auto-regressive Transformer model with a generative pre-training objective that predicts the user's next command given its prior commands in the sequence. We use a 12-layer Transformer which follows the architecture of the GPT-tiny model.
\item \textbf{util2vec~\cite{yang2017personalizing}}: We implement the method as described in the util2vec paper. In util2vec, the model is trained to predict the held-out central command given its surrounding commands and a context vector representing the user. The model weights are kept frozen after the training stage. In order to represent a user that is outside the training set and does not have a context vector during the testing time, a new context vector is optimized with the same objective over the user's command sequence. This context vector is then used as the user representation to make the final user-level predictions in the downstream tasks. 
\item \textbf{log2int~\cite{tao2019log2intent}}: We adapt the method described in log2int since our setting is not identical to theirs, which does not include paired descriptions of the command sequence. We keep the main sequence-to-sequence model of two recurrent neural networks, which are trained with an objective to predict commands in the next session given the current session. We then use the trained encoder network to generate the user's representation vector, which is passed to a linear classifier for the downstream tasks.
\end{itemize}
We use the same representation vector dimension of 256 for all methods to ensure fair comparison. Each method is repeated 3 times with different initialization to ensure stable performance.

\subsection{Main Results}
Table~\ref{tab:main} shows the main experimental results using full labeled subsets. We report the overall accuracy and F1-score for both downstream tasks as well as the overall performance. We directly use SciPy's weighted F1-score for multi-class classification, which better handles class imbalance. 
It can be seen that as we explained and expected in Section~\ref{sec:data}, the experience prediction Task 1 is more difficult than the expertise prediction Task 2. 
All learning-based methods produce results significantly better than the random baseline, which verifies the basic assumption that user-level information can be effectively extracted from their command sequences. 
The BoW baseline has the lowest overall accuracy among all learning-based methods as it is the most basic method. 
We find the CF-IUF method prone to favor dominant classes. It produces better overall accuracy, but worse F1-score especially in Task 1. This is due to the highly imbalanced nature of class distribution in the years of experience using the Fusion 360 software, whose user pool has been constantly increasing over time.
The GPT method is the second strongest method across both tasks and both metrics. It performs particularly well in Task 2, which is only one percentile below our proposed method. This is due to its large model capacity, which is good at capturing the theme which can be reflected at the command level. In the more difficult Task 1 that requires greater session- and user- level understanding, our model demonstrates significant advantages over GPT.
The util2vec method is trained using local command sequence windows of size $\pm 5$, which leads to reasonable performance in Task 2, but performs poorly in Task 1 that requires higher-level understanding.
The SeqAE and log2int methods are similar in network architecture, both using an encoder and decoder recurrent neural network. The main difference is SeqAE tries to reconstruct the same session, while log2int tries to predict the next session. In our dataset, adjacent sessions are often less related due to the rich functionality of Fusion 360 and diverse user behaviour. Therefore, log2int does not demonstrate a significant gain over SeqAE.
Our proposed \textit{SimCURL} method outperforms all comparisons in both tasks and metrics, demonstrating the effectiveness of its learned user representations.
In terms of efficiency, our method is lightweight and contains only $1.37$ million parameters, with an average inference time of $12.22$ milliseconds, making it suitable for large scale deployment and interactive applications.
For the experience prediction task, besides the default 8-class setting we merged 0-to-2, 2-to-5, and 5-to-8 year into a 3-class setting of beginner, intermediate, and experienced. Our method achieves 64.19\% accuracy and 59.86\% F1-score under this practical class setting.

\begin{table}[t!]
\caption{Ablation study with additional masked language modeling loss as well as without session dropout, self-supervised learning, user network branch, and session network branch.}
\vspace{-5mm}
\setlength{\tabcolsep}{6pt}
\begin{center}
\begin{tabular}{cccc|cc}
\hline
\textit{ses.dro.} & \textit{\,\,SSL\,} & \textit{\,\,user\,} & \textit{session} & Acc./\% & F1/\%\\\hline
\checkmark & \checkmark & \checkmark & \checkmark & \textbf{53.03} & \textbf{47.75}\\ 
\multicolumn{4}{c|}{\textit{+ MLM}} & 53.25(+0.22) & 47.82(+0.07)\\ \hline
$\times$ & $\times$ & \checkmark & \checkmark & 52.13(-0.90) & 46.40(-1.35)\\
\checkmark & $\times$ & \checkmark & \checkmark & 51.29(-2.01) & 46.43(-1.32)\\
\checkmark & \checkmark & $\times$ & \checkmark & 52.24(-0.79) & 46.09(-1.66)\\
\checkmark & \checkmark & \checkmark & $\times$ & 51.37(-1.30) & 46.76(-0.99)\\
\hline
\end{tabular}
\end{center}
\label{tab:ablation}
\end{table}

\begin{figure}
\includegraphics[width=\columnwidth]{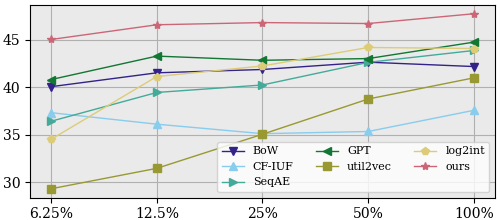}
\vspace{-7mm}
\caption{Few-shot learning: overall F1-score with varying amount of available supervision in the labeled subset.}
\label{fig:fewshot}
\end{figure}

\subsection{Few-Shot Learning}
We now evaluate our method in a few-shot learning setting that reflects a practical scenario where it is difficult to acquire large amount of labeled data. To simulate this, we train the models using only 6.25\% (1/16) of the training set. As shown in Table~\ref{tab:fewshot}, the util2vec method fails to work well under the few-shot setting, which performs only slightly better than random.
All methods suffer performance loss due to the scarcity of training data.
Light-weight models, especially CF-IUF, suffer the least compared to other models.
Our \textit{SimCURL} method shows strong few-shot learning ability, achieving best or second best performance in both tasks and overall. SimCURL also demonstrate significant advantage on the overall F1-score, outperforming the second best approach by more than 4\%.
We further show more in-depth few-shot learning results in Figure~\ref{fig:fewshot}, where the amount of available supervision decreases exponentially. It can be seen that our method has strong few-shot capability under scarce supervision as well as scalability as supervision increases.
Our method consistently outperforms the second-best method by more than 2.5\% across different supervision settings, where the improvement is the most significant at 4.22\% when supervision is the most scarce.

\begin{figure}[t!]
\setlength{\tabcolsep}{0pt}
\begin{center}
\begin{tabular}{c|cccc}
\hline
~ & {2L,512B} & {2L,1024B} & {2L,2048B} & {3L,512B}\\ \hline
\rotatebox[origin=l]{90}{Accuracy} &
\includegraphics[width=0.11\textwidth]{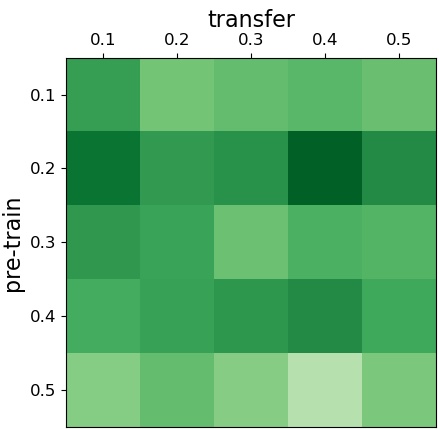} & 
\includegraphics[width=0.11\textwidth]{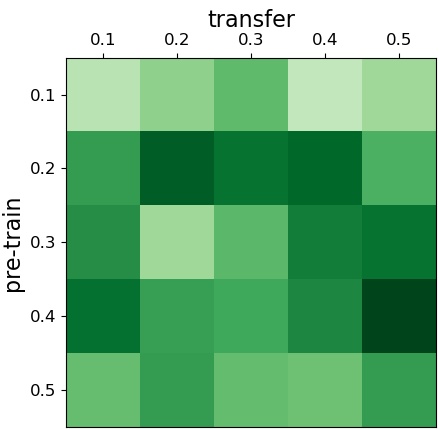} &
\includegraphics[width=0.11\textwidth]{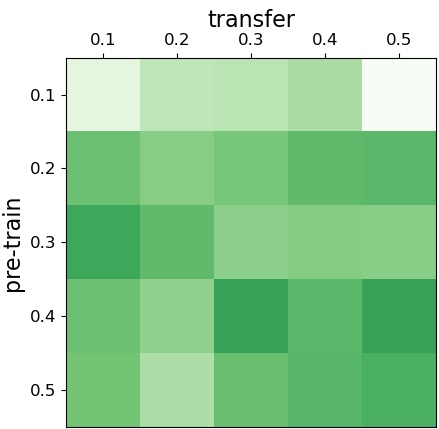} & 
\includegraphics[width=0.11\textwidth]{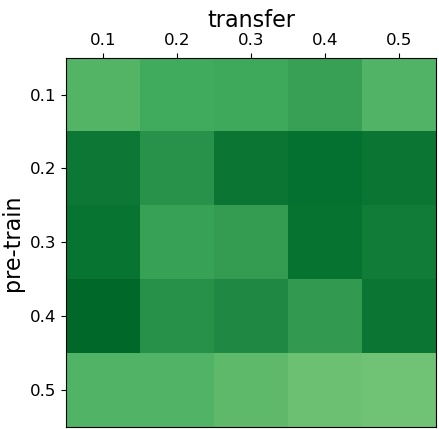} \\ \hline
\rotatebox[origin=l]{90}{F1-score} & 
\includegraphics[width=0.11\textwidth]{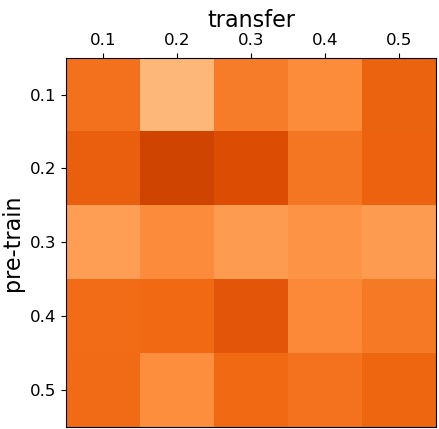} &
\includegraphics[width=0.11\textwidth]{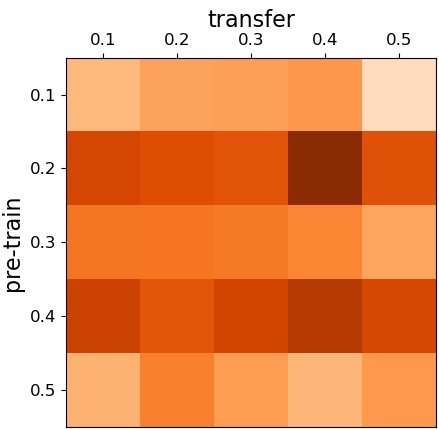} &
\includegraphics[width=0.11\textwidth]{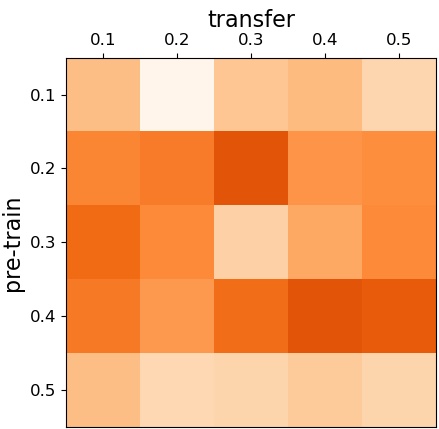} &
\includegraphics[width=0.11\textwidth]{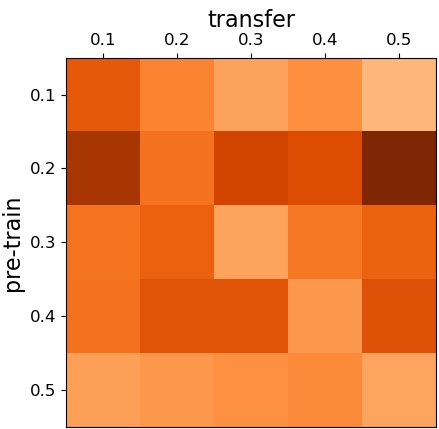}\\ \hline
\end{tabular}
\end{center}
\vspace{-3mm}
\caption{Effects of session dropout rate in the pre-training (y-axis) and transfer (x-axis) and stages, network depth (L), and pre-training batch size (B) in Task 1. Darker color means higher number: Acc.$\in$[36.15,37.87]/\%, F1$\in$[33.44,36.41]/\%.}
\label{fig:hyper}
\end{figure}

\subsection{Analysis}
To verify the effectiveness of different components of our method, we conduct an ablation study as shown in Table~\ref{tab:ablation}. We find the best performance is achieved with our full method, where the self-supervised learning brings the most gain in performance. The first ablation method that does not have session dropout also does not have the self-supervised learning stage, since the generation of views relies on session dropout. Interestingly, not using session dropout performs slightly better than using session dropout but without the self-supervised learning stage. This indicates the synergy between session dropout and self-supervised learning.
We also trained the representation model with an additional masked language model loss~\cite{devlin2018bert}, where sessions are tokenized with K-means clustering. This achieves minor improvement as shown in Table~\ref{tab:ablation}, while we leave further thorough study of its effect as future work.  
We study the effects of session dropout rate in the transfer and the self-supervised learning stages, under different network depths and batch sizes as shown in Figure~\ref{fig:hyper}. It can be seen that deeper network produces better performance, while there is no significant gain for batch size larger than 1024. The sweet spot for session dropout during self-supervised learning is between 0.2 and 0.4, while a larger rate during transfer learning usually helps.

\section{Conclusion}
The emerging data modality of command sequences possesses rich information for user modeling and personalization. 
In this paper we proposed a simple yet effective contrastive learning method named \textit{SimCURL}, that learns user representation vectors from unlabelled command stream records.
Our method is enabled by a user-session network architecture, and a new session dropout data augmentation technique.
We trained and evaluated on a real-world dataset of Fusion 360 command sequences consisting of more than half a billion commands.
Our method achieved significant improvement over existing methods on downstream classification tasks, which advanced user modeling techniques that can be used to enable better user experience.

\bibliographystyle{ieee_fullname}
\bibliography{main}

\begin{thebibliography}{10}\itemsep=-1pt

\bibitem{brown2020language}
Tom~B Brown, Benjamin Mann, Nick Ryder, Melanie Subbiah, Jared Kaplan, Prafulla
  Dhariwal, Arvind Neelakantan, Pranav Shyam, Girish Sastry, Amanda Askell,
  et~al.
\newblock Language models are few-shot learners.
\newblock {\em arXiv preprint arXiv:2005.14165}, 2020.

\bibitem{caron2021emerging}
Mathilde Caron, Hugo Touvron, Ishan Misra, Herv{\'e} J{\'e}gou, Julien Mairal,
  Piotr Bojanowski, and Armand Joulin.
\newblock Emerging properties in self-supervised vision transformers.
\newblock {\em arXiv preprint arXiv:2104.14294}, 2021.

\bibitem{chen2020simple}
Ting Chen, Simon Kornblith, Mohammad Norouzi, and Geoffrey Hinton.
\newblock A simple framework for contrastive learning of visual
  representations.
\newblock In {\em International conference on machine learning}, pages
  1597--1607. PMLR, 2020.

\bibitem{chung2014empirical}
Junyoung Chung, Caglar Gulcehre, KyungHyun Cho, and Yoshua Bengio.
\newblock Empirical evaluation of gated recurrent neural networks on sequence
  modeling.
\newblock {\em arXiv preprint arXiv:1412.3555}, 2014.

\bibitem{dai2015semi}
Andrew~M Dai and Quoc~V Le.
\newblock Semi-supervised sequence learning.
\newblock {\em Advances in neural information processing systems},
  28:3079--3087, 2015.

\bibitem{devlin2018bert}
Jacob Devlin, Ming-Wei Chang, Kenton Lee, and Kristina Toutanova.
\newblock Bert: Pre-training of deep bidirectional transformers for language
  understanding.
\newblock {\em arXiv preprint arXiv:1810.04805}, 2018.

\bibitem{fang2021ssml}
Xiaomin Fang, Jizhou Huang, Fan Wang, Lihang Liu, Yibo Sun, and Haifeng Wang.
\newblock Ssml: Self-supervised meta-learner for en route travel time
  estimation at baidu maps.
\newblock In {\em Proceedings of the 27th ACM SIGKDD Conference on Knowledge
  Discovery \& Data Mining}, pages 2840--2848, 2021.

\bibitem{gao2021simcse}
Tianyu Gao, Xingcheng Yao, and Danqi Chen.
\newblock Simcse: Simple contrastive learning of sentence embeddings.
\newblock {\em arXiv preprint arXiv:2104.08821}, 2021.

\bibitem{gao2022command}
Wen Gao, Xuanming Zhang, Qiushi He, Borong Lin, and Weixin Huang.
\newblock Command prediction based on early 3d modeling design logs by deep
  neural networks.
\newblock {\em Automation in Construction}, 133:104026, 2022.

\bibitem{grill2020bootstrap}
Jean-Bastien Grill, Florian Strub, Florent Altch{\'e}, Corentin Tallec,
  Pierre~H Richemond, Elena Buchatskaya, Carl Doersch, Bernardo~Avila Pires,
  Zhaohan~Daniel Guo, Mohammad~Gheshlaghi Azar, et~al.
\newblock Bootstrap your own latent: A new approach to self-supervised
  learning.
\newblock {\em arXiv preprint arXiv:2006.07733}, 2020.

\bibitem{hadsell2006dimensionality}
Raia Hadsell, Sumit Chopra, and Yann LeCun.
\newblock Dimensionality reduction by learning an invariant mapping.
\newblock In {\em 2006 IEEE Computer Society Conference on Computer Vision and
  Pattern Recognition (CVPR'06)}, volume~2, pages 1735--1742. IEEE, 2006.

\bibitem{he2020momentum}
Kaiming He, Haoqi Fan, Yuxin Wu, Saining Xie, and Ross Girshick.
\newblock Momentum contrast for unsupervised visual representation learning.
\newblock In {\em Proceedings of the IEEE/CVF Conference on Computer Vision and
  Pattern Recognition}, pages 9729--9738, 2020.

\bibitem{he2020loghub}
Shilin He, Jieming Zhu, Pinjia He, and Michael~R Lyu.
\newblock Loghub: A large collection of system log datasets towards automated
  log analytics.
\newblock {\em arXiv preprint arXiv:2008.06448}, 2020.

\bibitem{hochreiter1997long}
Sepp Hochreiter and J{\"u}rgen Schmidhuber.
\newblock Long short-term memory.
\newblock {\em Neural computation}, 9(8):1735--1780, 1997.

\bibitem{kingma2014adam}
Diederik~P Kingma and Jimmy Ba.
\newblock Adam: A method for stochastic optimization.
\newblock {\em arXiv preprint arXiv:1412.6980}, 2014.

\bibitem{lecun2021self}
Y LeCun and I Misra.
\newblock Self-supervised learning: The dark matter of intelligence, 2021.

\bibitem{liu2019roberta}
Yinhan Liu, Myle Ott, Naman Goyal, Jingfei Du, Mandar Joshi, Danqi Chen, Omer
  Levy, Mike Lewis, Luke Zettlemoyer, and Veselin Stoyanov.
\newblock Roberta: A robustly optimized bert pretraining approach.
\newblock {\em arXiv preprint arXiv:1907.11692}, 2019.

\bibitem{matejka2009communitycommands}
Justin Matejka, Wei Li, Tovi Grossman, and George Fitzmaurice.
\newblock Communitycommands: command recommendations for software applications.
\newblock In {\em Proceedings of the 22nd annual ACM symposium on User
  interface software and technology}, pages 193--202, 2009.

\bibitem{mei2016neural}
Hongyuan Mei and Jason Eisner.
\newblock The neural hawkes process: A neurally self-modulating multivariate
  point process.
\newblock {\em arXiv preprint arXiv:1612.09328}, 2016.

\bibitem{mikolov2013efficient}
Tomas Mikolov, Kai Chen, Greg Corrado, and Jeffrey Dean.
\newblock Efficient estimation of word representations in vector space.
\newblock {\em arXiv preprint arXiv:1301.3781}, 2013.

\bibitem{niizumi2021byol}
Daisuke Niizumi, Daiki Takeuchi, Yasunori Ohishi, Noboru Harada, and Kunio
  Kashino.
\newblock Byol for audio: Self-supervised learning for general-purpose audio
  representation.
\newblock {\em arXiv preprint arXiv:2103.06695}, 2021.

\bibitem{pennington2014glove}
Jeffrey Pennington, Richard Socher, and Christopher~D Manning.
\newblock Glove: Global vectors for word representation.
\newblock In {\em Proceedings of the 2014 conference on empirical methods in
  natural language processing (EMNLP)}, pages 1532--1543, 2014.

\bibitem{rabiner1989tutorial}
Lawrence~R Rabiner.
\newblock A tutorial on hidden markov models and selected applications in
  speech recognition.
\newblock {\em Proceedings of the IEEE}, 77(2):257--286, 1989.

\bibitem{radford2018improving}
Alec Radford, Karthik Narasimhan, Tim Salimans, Ilya Sutskever, et~al.
\newblock Improving language understanding by generative pre-training.
\newblock {\em OpenAI}, 2018.

\bibitem{radford2019language}
Alec Radford, Jeffrey Wu, Rewon Child, David Luan, Dario Amodei, Ilya
  Sutskever, et~al.
\newblock Language models are unsupervised multitask learners.
\newblock {\em OpenAI blog}, 1(8):9, 2019.

\bibitem{raffel2019exploring}
Colin Raffel, Noam Shazeer, Adam Roberts, Katherine Lee, Sharan Narang, Michael
  Matena, Yanqi Zhou, Wei Li, and Peter~J Liu.
\newblock Exploring the limits of transfer learning with a unified text-to-text
  transformer.
\newblock {\em arXiv preprint arXiv:1910.10683}, 2019.

\bibitem{schneider2019wav2vec}
Steffen Schneider, Alexei Baevski, Ronan Collobert, and Michael Auli.
\newblock wav2vec: Unsupervised pre-training for speech recognition.
\newblock {\em arXiv preprint arXiv:1904.05862}, 2019.

\bibitem{tao2019log2intent}
Zhiqiang Tao, Sheng Li, Zhaowen Wang, Chen Fang, Longqi Yang, Handong Zhao, and
  Yun Fu.
\newblock Log2intent: Towards interpretable user modeling via recurrent
  semantics memory unit.
\newblock In {\em Proceedings of the 25th ACM SIGKDD International Conference
  on Knowledge Discovery \& Data Mining}, pages 1055--1063, 2019.

\bibitem{vaswani2017attention}
Ashish Vaswani, Noam Shazeer, Niki Parmar, Jakob Uszkoreit, Llion Jones,
  Aidan~N Gomez, {\L}ukasz Kaiser, and Illia Polosukhin.
\newblock Attention is all you need.
\newblock In {\em Advances in neural information processing systems}, pages
  5998--6008, 2017.

\bibitem{viterbi1967error}
Andrew Viterbi.
\newblock Error bounds for convolutional codes and an asymptotically optimum
  decoding algorithm.
\newblock {\em IEEE transactions on Information Theory}, 13(2):260--269, 1967.

\bibitem{wang2020beyond}
Wen Wang, Wei Zhang, Shukai Liu, Qi Liu, Bo Zhang, Leyu Lin, and Hongyuan Zha.
\newblock Beyond clicks: Modeling multi-relational item graph for session-based
  target behavior prediction.
\newblock In {\em Proceedings of The Web Conference 2020}, pages 3056--3062,
  2020.

\bibitem{wang2018leveraging}
Xu Wang, Benjamin Lafreniere, and Tovi Grossman.
\newblock Leveraging community-generated videos and command logs to classify
  and recommend software workflows.
\newblock In {\em Proceedings of the 2018 CHI Conference on Human Factors in
  Computing Systems}, pages 1--13, 2018.

\bibitem{xiong2018session}
Wayne Xiong, Lingfeng Wu, Jun Zhang, and Andreas Stolcke.
\newblock Session-level language modeling for conversational speech.
\newblock In {\em Proceedings of the 2018 Conference on Empirical Methods in
  Natural Language Processing}, pages 2764--2768, 2018.

\bibitem{xue2021exploring}
Hao Xue and Flora~D Salim.
\newblock Exploring self-supervised representation ensembles for covid-19 cough
  classification.
\newblock In {\em Proceedings of the 27th ACM SIGKDD Conference on Knowledge
  Discovery \& Data Mining}, pages 1944--1952, 2021.

\bibitem{yang2019creative}
Longqi Yang, Chen Fang, Hailin Jin, Walter Chang, and Deborah Estrin.
\newblock Creative procedural-knowledge extraction from web design tutorials.
\newblock {\em arXiv preprint arXiv:1904.08587}, 2019.

\bibitem{yang2017personalizing}
Longqi Yang, Chen Fang, Hailin Jin, Matthew~D Hoffman, and Deborah Estrin.
\newblock Personalizing software and web services by integrating unstructured
  application usage traces.
\newblock In {\em Proceedings of the 26th International Conference on World
  Wide Web Companion}, pages 485--493, 2017.

\bibitem{yang2018characterizing}
Longqi Yang, Chen Fang, Hailin Jin, Matthew~D Hoffman, and Deborah Estrin.
\newblock Characterizing user skills from application usage traces with
  hierarchical attention recurrent networks.
\newblock {\em ACM Transactions on Intelligent Systems and Technology (TIST)},
  9(6):1--18, 2018.

\end{thebibliography}

\end{document}